\documentclass[sigconf]{acmart}
\AtBeginDocument{%
  }

\setcopyright{acmlicensed}
\copyrightyear{xxxx}
\acmYear{xxxx}
\acmDOI{XXXXXXX.XXXXXXX}
\acmConference[Conference acronym 'XX]{Make sure to enter the correct
  conference title from your rights confirmation email}{June 03--05,
  xxxx}{Woodstock, NY}
\acmISBN{978-1-4503-XXXX-X/2018/06}

\usepackage{graphicx}
\usepackage{amsfonts}
\usepackage{amsmath}
\usepackage{multirow}
\usepackage{xcolor}
\usepackage{booktabs}
\usepackage{algorithm}
\usepackage{algpseudocode}
\usepackage{setspace}

\usepackage{pifont}
\newcommand{\cmark}{\ding{51}}
\newcommand{\xmark}{\ding{55}}

\begin{document}

\title{From Semantics, Scene to Instance-awareness: Distilling Foundation Model for Open-vocabulary \\ Grounded Situation Recognition}


\author{Chen Cai}
\affiliation{%
  \institution{National University of Singapore,}
  \country{Singapore}
}

\author{Tianyi Liu}
\affiliation{%
  \institution{Nanyang Technological University,}
  \country{Singapore}}

\author{Jianjun Gao}
\affiliation{%
  \institution{Nanyang Technological University,}
  \country{Singapore}}

\author{Wenyang Liu}
\affiliation{%
  \institution{Nanyang Technological University,}
  \country{Singapore}}

\author{Kejun Wu}
\authornote{Corresponding author. Chen Cai: E190210@e.ntu.edu.sg; Kejun Wu: kjwu@hust.edu.sg, Yi Wang: yi-eie.wang@polyu.edu.hk}
\affiliation{%
  \country{Huazhong University of Science and Technology, China}}

\author{Ruoyu Wang}
\affiliation{%
  \institution{Nanyang Technological University,}
  \country{Singapore}}

\author{Yi Wang\footnotemark[1]}
\affiliation{%
  \country{The Hong Kong Polytechnic University, Hong Kong SAR}}

\author{Soo Chin Liew}
\affiliation{%
  \country{National University of Singapore, Singapore}
  }

\renewcommand{\shortauthors}{Trovato et al.}

\begin{abstract}
Recent Multimodal Large Language Models (MLLMs) exhibit strong zero-shot abilities but struggle with complex Grounded Situation Recognition (GSR) and are resource-intensive for edge device deployment.
Meanwhile, conventional GSR models often lack generalization ability, falling short in recognizing unseen and rare situations.
In this paper, we exploit transferring knowledge from a teacher MLLM to a small GSR model to enhance its generalization and zero-shot abilities, thereby introducing the task of Open-vocabulary Grounded Situation Recognition (Ov-GSR).
To achieve this, we propose Multimodal Interactive Prompt Distillation (MIPD), a novel framework that distills enriched multimodal knowledge from the foundation model, enabling the student Ov-GSR model to recognize unseen situations and be better aware of rare situations.
Specifically, the MIPD framework first leverages the LLM-based Judgmental Rationales Generator (JRG) to construct positive and negative glimpse and gaze rationales enriched with contextual semantic information. 
The proposed scene-aware and instance-perception prompts are then introduced to align rationales with visual information from the MLLM teacher via the Negative-Guided Multimodal Prompting Alignment (NMPA) module, effectively capturing holistic and perceptual multimodal knowledge.
Finally, the aligned multimodal knowledge is distilled into the student Ov-GSR model, providing a stronger foundation for generalization that enhances situation understanding, bridges the gap between seen and unseen scenarios, and mitigates prediction bias in rare cases.
We evaluate MIPD on the refined Ov-SWiG dataset, achieving superior performance on seen, rare, and unseen situations, and further demonstrate improved unseen detection on the HICO-DET dataset. Data and code: https://github.com/caicch/Ov-GSR
\end{abstract}





\maketitle

\begin{figure}[t!]
\centering
  \centerline{\includegraphics[width=8cm]{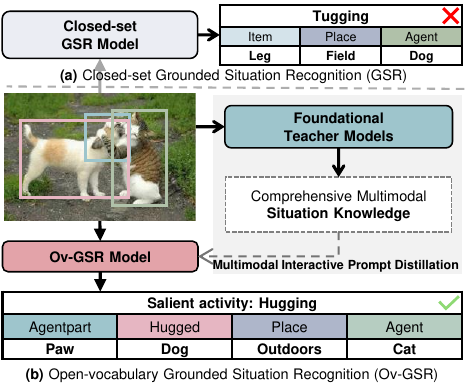}} \vspace{-14pt}
  \caption{Illustration of conventional closed-set Grounded Situation Recognition (GSR) and the proposed Open-vocabulary GSR (Ov-GSR). (a) Closed-set GSR methods fail to predict unseen activities (e.g., Hugging), resulting in incorrect semantic role recognition. (b) Ov-GSR gains the ability to identify unseen situations through the proposed Multimodal Interactive Prompt Distillation (MIPD) framework. For example, it correctly predicts the salient activity “Hugging,” along with its semantic roles and detects the entities: AgentPart: Paw, Hugged: Dog, Place: Outdoors, and Agent: Cat.}
   \label{fig:1} 
   \vspace{-10pt}
\end{figure}


\section{Introduction}
\label{sec:intro}

\begin{figure}[t!]
\centering
  \includegraphics[width=8cm, height=5.2cm]{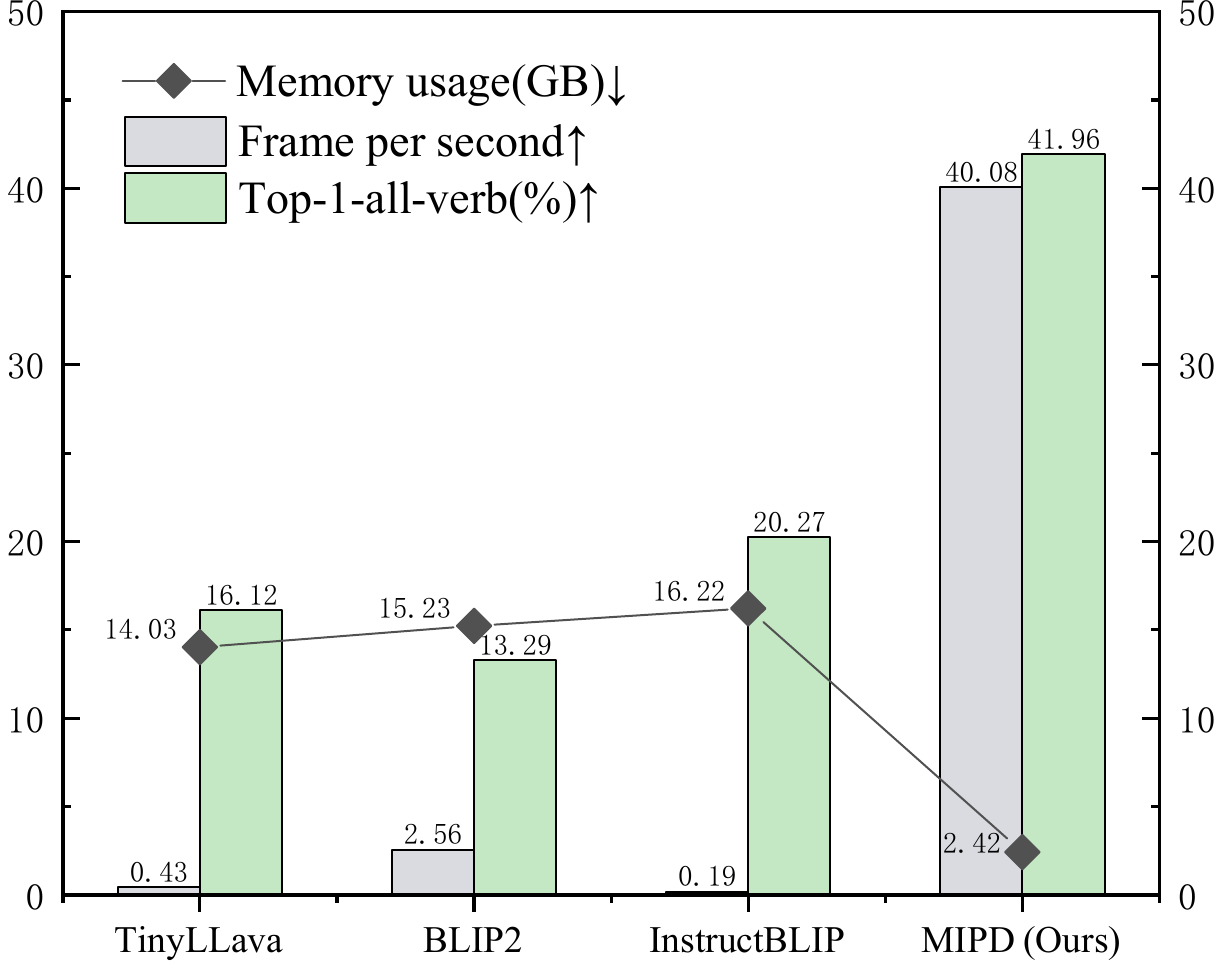} \vspace{-12pt}
  \caption{The analysis compares inference resource requirements with existing larger models. Ours uses lower memory for deployment and has faster Frame Per Second.}
  \label{fig:tab9} 
  \vspace{-10pt}
\end{figure}

The ability to conduct situation recognition in the scene is one of the essential roles of vision~\cite{instructblip, Bitstream, TimeAction, motionManba} and language~\cite{llama2, blip2} research, with broad applications in assistive technology such as autonomous driving~\cite{vlp, vlm2scene, wang2024cm} and visual impairments~\cite{vialm}.
Recently, Multimodal Large Language Models (MLLMs)~\cite{instructblip, minigpt} have demonstrated remarkable zero-shot scene understanding and can be applied across various domains~\cite{CL-HOI, Colpro, DisVLM_OOD}. 
However, many of them rely on smaller large language model (LLM) counterparts (e.g., InstructBLIP~\cite{instructblip}, TinyLLaVA~\cite{llava}), which often underperform in tasks like Grounded Situation Recognition (GSR)~\cite{GSR, VisualActivity} that require deep comprehension~\cite{emergent1, emergent2}, as reflected by the Top-1 activity prediction accuracy (Top-1-all-verb) in Figure~\ref{fig:tab9}.
Moreover, although these models are relatively small compared to much larger ones (e.g., with 34B~\cite{yi01,llama2} parameters), fine-tuning and deploying such massive models for GSR remain challenging due to their substantial computational and resource demands. 
This issue is particularly critical for apply GSR to many assistive technologies, which often depend on \textit{low-resource edge devices} rather than heavy servers with modern GPUs~\cite{TGK, efficientCap, lightweightOD}. 
Addressing this problem is crucial for advancing the development of small and efficient GSR model capable of accurately interpreting complex scenes while preserving the generalization capabilities of MLLMs, potentially \textit{benefiting a wide range of assistive technologies}.

In this paper, we exploit distilling the scene interpretation capabilities of large model into a small and efficient GSR model, which summarizes complex scenes by identifying what is happening (activity), who or what is involved (entities), and where they are located (coordinates), as illustrated in Figure~\ref{fig:1}.
Existing GSR methods~\cite{gsrformer, coformer} aim to identify hundreds of activities along with their corresponding entities across various situations. While these state-of-the-art methods demonstrate remarkable performance, they face two major challenges: (\textbf{C1}) limited to predicting visual concepts within predefined seen situations~\cite{CMD-SE, UnseenVCS}. In real-world scenarios, a GSR model is highly likely to encounter situations from unseen categories that were not present in the training data. The recognition abilities of these conventional GSR models degrade when inferring over unseen scenario. (\textbf{C2}) struggling to recognize rare situations due to data imbalance.
Within the dataset~\cite{GSR}, some situations are abundantly represented, while others have fewer samples, which cause the model to attend more on
the frequently appeared situations and tends to miss recognizing the rare situations.
These challenges motivate us to develop a method that distills the generalization and robust scene understanding capabilities of the large teacher model into a smaller GSR model, enabling more effective recognition of unseen and rare situations while supporting deployment on edge devices (e.g., Figure~\ref{fig:tab9} MIPD (Ours)).
Formally, we define a new problem setting (Sec.~\ref{Problem_Overview}) as \textbf{Open-vocabulary Grounded Situation Recognition (Ov-GSR)}, as illustrated in Figure~\ref{fig:1} (b). 

To this end, we propose the Multimodal Interactive Prompt Distillation (MIPD) framework, a novel approach that distills multimodal knowledge from the large teacher model to enhance the generalized recognition abilities of the smaller student Ov-GSR model, improving its capacity to better understand seen, rare and unseen situations.
Specifically, MIPD integrates rich contextual semantic knowledge generated by the Judgmental Rationales Generator (JRG) along with scene-aware and instance-perception information from the MLLM, providing a comprehensive and diverse knowledge foundation for the student model to learn from.
First, we use the MLLM with the JRG to generate reliable positive and negative glimpse and gaze rationales through LLM-based judgment~\cite{judging, judgesurvey} and multi-round reasoning.
These rationales enhance the student model's semantic understanding by integrating both glimpse and gaze-level insights, bridging the knowledge gap between seen and unseen situations (for \textbf{C1}) and fostering knowledge to mitigate imbalanced predictions for rare scenarios (for \textbf{C2}), ultimately benefiting open-vocabulary situation understanding.
Furthermore, the introduced learnable scene-aware and instance-perception prompts are designed to capture rich scene-level and regional entity-level visual knowledge from the teacher MLLM.
These prompts are aligned with glimpse and gaze rationales through the Negative-Guided Multimodal Prompting Alignment (NMPA) module, effectively integrating and distilling both holistic and perception-level multimodal knowledge into the student model.
Through the distillation process, the knowledge aligned in the prompts enhances the student model’s understanding of activities and entities for Ov-GSR, improving its ability to recognize both rare (\textbf{C2}) and unseen (\textbf{C1}) situations.
With MIPD, the student model encapsulates rich multimodal information from the large model for better Ov-GSR performance.

Our main contributions are summaries as: (1) We explore the novel problem of Open-vocabulary Grounded Situation Recognition (Ov-GSR) and highlight a critical challenge: enabling small models to develop generalization capabilities for recognizing unseen and rare situations. This challenge motivates our empirical investigation to effectively address the problem. (2) We propose the Multimodal Interactive Prompt Distillation (MIPD) framework, which leverages glimpse and gaze rationales enriched with semantic information, aligned with scene-aware and instance-perception prompts, to effectively transfer multimodal knowledge from the teacher large model to a student Ov-GSR model. 
(3) We evaluate Ov-GSR performance through extensive experiments on the newly split Ov-SWiG dataset, covering seen, rare, and unseen situations.
It shows that MIPD achieves state-of-the-art results. Besides, we apply MIPD to Human-Object Interaction (HOI) detection and improve performance on unseen detection in the HICO-DET dataset.


\section{Related Works}


\subsection{Grounded Situation Recognition}

Grounded Situation Recognition (GSR)~\cite{GSR, coformer,  SBC-GSR, VisualActivity, dynamicSR} is a fundamental scene understanding task that involves identifying activities, detecting relevant roles with their corresponding entities and bounding boxes. It has a wide range of real-world applications for edge assistive technologies, such as visual impairment and autonomous driving systems.
Existing methods have made significant progress in improving small closed-set GSR. 
CoFormer~\cite{coformer} proposes a collaborative transformer that jointly leverages multiple transformers for activity prediction and entity detection.
OpenSU~\cite{openSU} enhances GSR by enabling dense segmentation through the integration of the segment anything Model~\cite{sam} as a segmentation mask generator, leading to improved scene comprehension.
ClipSitu~\cite{clipsitu} strengthens activity and entities recognition by incorporating the CLIP foundational vision-language model for more comprehensive situational awareness.
Existing methods focus on predicting closed-set situations, which limits the model's ability to recognize unseen situations. In this paper, we explore an Ov-GSR model that enhances the recognition of unseen and rare situations by combining rich knowledge from large models.


\subsection{Knowledge Distillation of Large Models}


Recent advanced works distill large model capabilities into smaller ones~\cite{freekd, PRR, Dime-fm, promptmm}, demonstrating promising results. 
Minmax~\cite{Minmax} formulated dataset distillation as a minmax optimization problem and proposed neural characteristic function discrepancy to effectively measure distributional differences, enabling compact and high-quality synthetic dataset generation.
PRR~\cite{PRR} introduces a retrieval-based Chain-of-Thought (CoT)~\cite{cot} distillation technique, which transfer knowledge from LLMs to smaller language models, enhancing the performance of the question answering tasks.
Tinyllm~\cite{tinyllm} introduces a new knowledge distillation paradigm, where a small student LLM learns from multiple large teacher models, effectively capturing knowledge from multiple rationals while maintaining efficiency. 
Some methods focus on knowledge distillation through efficient prompt-tuning~\cite{visualPT, promptsr, visionprompt, visionprompt2}. PromptMM~\cite{promptmm} enhances recommender systems by leveraging prompt-tuning, enabling efficient distillation to bridge the semantic gap across multi-modal contexts. PromptKD~\cite{promptkd} leverages soft prompt-based imitation on unlabeled domain images, allowing a lightweight target model to acquire knowledge from a large teacher model through a novel unsupervised distillation approach. Differently, this work encapsulates the information of rich rationales and visual prompts from the teacher model to effectively distill multimodal knowledge into a student model for improved Ov-GSR.

\vspace{-8pt}

\subsection{Open-vocabulary Tasks}

\vspace{-2pt}

Recent research has focused on transferring the open-vocabulary capabilities of MLLMs to downstream tasks such as object detection~\cite{Ovdet_VLM, contextualdet}, human-object interaction~\cite{UniversalHoi, THID}, and image classification~\cite{ov_cls1, ovmr}. OVMR~\cite{ovmr} leverages multimodal cues, combining textual descriptions and exemplar images to facilitate recognition. 
ContextDET~\cite{contextualdet} introduces a unified multimodal model that integrates an LLM to learn visual-language contexts, enabling the model to identify and associate visual objects.

The most similar work to Ov-GSR is end-to-end open-vocabulary HOI detection~\cite{THID, CMD-SE}, which simultaneously recognizes actions in an image while detecting humans and objects. 
THID~\cite{THID} distills and utilizes transferable knowledge from the pretrained CLIP model, integrating multimodal features into a joint visual-text space to enhance open-vocabulary interaction detection.
CMD-SE~\cite{CMD-SE} presents a novel HOI detection framework that distills fine-grained human body part semantic knowledge from LLM to enhance interaction recognition. 
The proposed Ov-GSR focuses on recognizing the activity first, followed by the detection of multiple entities within an image.
In contrast to CMD-SE, our model adopts a different approach by aligning multimodal knowledge from semantics, scene to instance-level, effectively distilling this knowledge to bridge the gap between seen and unseen situations while enhancing rare situation awareness.

\begin{figure*}[t]
\centering
  \centerline{\includegraphics[width=18cm]{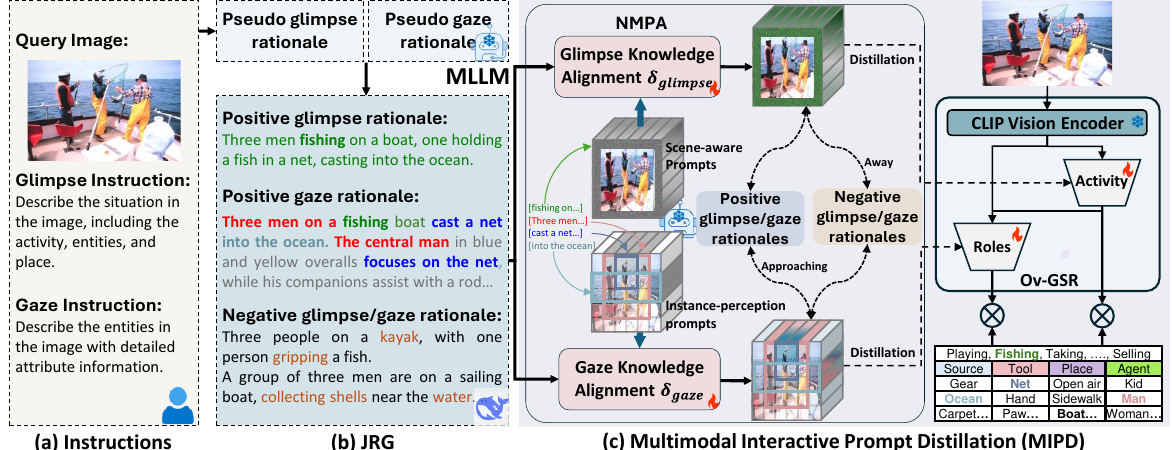}}\vspace{-8pt}
  \caption{Overview of our framework: We first leverage an MLLM guided with (a) instructions to generate pseudo glimpse and gaze rationales for scene and entity understanding. This is followed by the (b) Judgmental Rationales Generator (JRG), which employs an LLM-judge to evaluate and iteratively refine these rationales through multi-round reasoning, resulting in high-quality positive and negative rationales. 
These rationales are then aligned with scene-aware and instance-perception prompts to encapsulate visual and semantic information from teacher MLLM model through the Negative-Guided Multimodal Prompting Alignment (NMPA) module.
Finally, our proposed (c) Multimodal Interactive Prompt Distillation (MIPD) framework distills the aligned multimodal knowledge into the student model, enabling more accurate and generalizable Ov-GSR.}
  \label{fig:overview}
\end{figure*}

\section{Methodology}


\subsection{Problem Overview and Motivation}\label{Problem_Overview} 


In this section, we first present the problem overview, followed by our motivation for leveraging prompting and distillation strategies with the large models to achieve our goal.

\textbf{Grounded Situation Recognition (GSR):} aims to summarize visual content by analyzing \textit{what} is happening (activity understanding), \textit{who} and  \textit{what} are involved and their roles (entities recognition), and \textit{where} the entities are located (bounding box prediction). 
The introduced \textbf{Open-vocabulary Grounded Situation Recognition} (\textbf{Ov-GSR}) represents a more challenging problem, as it operates in a more generalized scenario. Specifically, the task involves training on a predefined set of base situations while extending the model's capability to predict \underline{unseen situations}.

Formally, let us define a set of base situation categories as $s^b = \{v^b, \mathcal{F}_v^b\} \in \mathcal{S}^b$, where $v^b \in \mathcal{V}^b$  represents the base salient activity, and its corresponding semantic roles are given by $\mathcal{F}_v^b = \{\mathbf{f_r} |\mathbf{f_r} = (r, n_r, c_r), \forall r \in \mathcal{R}_v, n_r \in \mathcal{N}^b, c_r \in \mathbb{R}^4\}$. Here, $r$ denotes the semantic role, $n_r$  represents the corresponding entity, and $c_r$ refers to the bounding box coordinates of that entity.
For instance, as shown in Figure~\ref{fig:1}, $\mathcal{F}_{hugging} = \{\mathbf{f_{agentpart}}, \mathbf{f_{hugged}}, \mathbf{f_{place}}, \mathbf{f_{agent}}\}$, where each role ($\mathbf{f_{agentpart}}$) contains respective entity and bounding box.
To align the complexity of a realistic open-world scenario, we assume the existence of \textit{unseen} situation categories $\mathcal{S}^u$, where $\mathcal{S}^u \cap \mathcal{S}^b = \emptyset$, containing \textit{novel situation} that are absent from the base set $\mathcal{S}^b$. 
The objective of Ov-GSR is to train a model using the base training set $\mathcal{D}^b=\{(x_i, s_i^b)\}_{i=1}^N$, where $N$ represents the total number of training images. Here, $x_i$ denotes the $i$-th image, and $s_i$ corresponds to its label, which includes the annotated situation category $s^b_i$.
During the inference stage, the model can predict situation of $\mathcal{S}=\mathcal{S}^b \cup \mathcal{S}^u$ with the unseen test set $\mathcal{D}=\mathcal{D}^b\cup\{(x_i, s_i^u)\}_{i=1}^M$, where $M$ denotes the number of unseen samples.

\textbf{Distill Knowledge from Large Models with Prompts:}
Knowledge distillation~\cite{distillsurvey, minillm, hoiclip} has emerged as a key technique to alleviate the substantial computational demands of modern MLLMs by training smaller student models to replicate the behavior of larger teacher models, significantly reducing resource consumption. Furthermore, the soft prompting technique~\cite{llama-Adv2, visualPT} has demonstrated advancements in efficiently fine-tuning models, allowing them to achieve strong performance with language instruction~\cite{m2pt, IAPT}, enabling effective execution of downstream tasks.

These motivate us to distill knowledge using efficient multimodal prompting techniques from the frozen teacher model $T_{\text{model}}(\cdot;\theta_T)$, parameterized by  $\theta_T$, which has been pre-trained on a large multimodality corpus, into the student model $S_{\text{model}}(\cdot;\theta_S)$, parameterized by $\theta_S$. This process enables the student model to inherit the strong capabilities of the teacher. The objective function is defined as $\mathcal{L} = \ell(S_{\text{model}}, T_{\text{model}})$, where $\ell$ denotes the objective function, such as KL divergence, L1 loss, or cross-entropy loss, computed between the learned output features of the student model, the teacher model, or the target output produced by the teacher.

\vspace{-10pt}

\subsection{Prompt Distillation Framework}

\vspace{-2pt}

Our proposed method is illustrated in Figure~\ref{fig:overview}. We introduce the Multimodal Interactive Prompt Distillation (MIPD) framework, which distills semantic, scene, and instance prompts knowledge from teacher MLLM to strengthen a student Ov-GSR model. This approach improves the model’s generalization capabilities, enabling it to effectively understand seen, rare, and unseen situations. 
In the MIPD process, scene-aware and instance-perception prompts are employed to align and integrate the glimpse and gaze rationales, which contain rich semantic information generated by the LLM based Judgmental Rationales Generator (JRG), with visual features extracted from the MLLM. 
Then, the aligned knowledge with prompts is distilled into the student model, enhancing its ability to recognize complex situations.
This alignment is achieved through the Negative-Guided Multimodal Prompting Alignment (NMPA) module, facilitating effective semantic and visual integration. 
The glimpse and gaze rationales serve as hard prompts, while scene-aware and instance-perception prompts function as learnable soft prompts, denoted as $\mathbf{P}_{gli}$, $\mathbf{P}_{gaz}$, $\mathbf{P}_{sce}$, and $\mathbf{P}_{ins}$, respectively. The distillation process can be formulated as:
\begin{equation}
\begin{aligned}
\theta_S^* = \arg\min_{\theta_S} \mathbb{E}_{(I, s) \sim \mathcal{D}} \Big[ 
& \ell(S_{\text{model}}(I; \theta_S), T_{\text{model}}(I, \mathbf{P}; \theta_T)) \\
& + \ell(S_{\text{model}}(I; \theta_S), s) \Big]
\end{aligned}
\end{equation}
where $\theta_S^*$ denotes the optimal parameters of the student model, $I, s, \mathbf{P}$ are the input image, situation label, and the prompts, respectively.

Given an input image $I$, the frozen vision encoder in the teacher MLLM network first extracts the visual feature maps  $\mathbf{X}_T = T_{model}(I)$, where $\mathbf{X}_T \in \mathbb{R} ^{H \times W \times D}$. 
Then, the prompts $\mathbf{P}_{sce}$ and $\mathbf{P}_{ins}$ are attached to $\mathbf{X}_T$ and interact with $\mathbf{P}_{gli}$ and $\mathbf{P}_{gaz}$ to model rich semantic, scene, and instance-level multimodal knowledge. 
This enriched knowledge enables the student model, $\mathbf{X}_S = S_{\text{model}}(I)$, to achieve improved Ov-GSR by enhancing its generalization through multimodal information distillation, bridging the gap between seen and unseen situations and reducing prediction bias in the rare scenario.


\vspace{-4pt}

\subsubsection{Rationales Generation with MLLM and LLM-judgment}
Excellent rationales, serving as contextual semantic information, have been shown in many recent studies to enhance model learning~\cite{CMD-SE, deepseek-r1, MMCOT, ddcot}. 
In this study, we distill richer semantic knowledge from reliable rationales generated by large models during training to improve Ov-GSR performance, enabling better recognition of rare and unseen situations. This process eliminates rationales at inference, enhancing model efficiency and deployment ability.

To generate high-quality situation-aware rationales, we first employ an MLLM to produce pseudo rationales enriched with visual information. We then integrate a Judgmental Rationales Generator (JRG), which incorporates a powerful language model (e.g., DeepSeek~\cite{deepseek-r1}, Gemini~\cite{gemini1.5}) as a judge, refining the rationales for improved coherence. 
During scene situation awareness, similar to how humans first cast a quick glimpse to understand what is happening before gradually gazing at details to identify involved objects and their relationships, the GSR model~\cite{GSR,coformer} initially comprehend the overall activity before focusing on detailed analysis to interpret the entities and their interactions within the situation.
Hence, we utilize JRG to generate glimpse-level rationales $\mathbf{P}_{gli}$ to facilitate overall scene activity understanding, followed by the generation of gaze-level rationales $\mathbf{P}_{gaz}$ to benefit in detailed entity comprehension within the scene. This can be formulated as:
\begin{equation}
    \mathbf{P}_{gli+}, \mathbf{P}_{gaz+}, \mathbf{P}_{gli-}, \mathbf{P}_{gaz-} = \text{JRG}(I, Intructions)
\end{equation}
here, we assume the expression of rationales $\mathbf{P} \in \mathbb{R}^{L \times D}$ are already encoded with a text encoder~\cite{clip} to ease the presentation, where $L$ denotes the length of the rationale and $D$ is the dimension.

More specifically, as illustrated in figure~\ref{fig:overview}, give a input image $I$, 
we first use an MLLM with instruction to generate general pseudo-glimpse and gaze rationales for scene and detailed entity understanding. 
While these rationales often struggle to accurately capture the expected visual situations~\cite{emergent1, emergent2}, we observed that they provided rich contextual semantic attributes such as color, pattern, and material, which can effectively support scene understanding~\cite{MMCOT, devilOv-detect} (see supplementary for examples).
Hence, to improve the accuracy and coherence of the rationals, we retain these meaningful attributes information while refining incorrect situation knowledge during the rationale generation process within JRG.

\begingroup
\setlength{\textfloatsep}{8pt}
\begin{algorithm}[t]
\caption{Judgmental Rationales Generator (JRG)}
\label{alg:llm-judge}
\begin{algorithmic}[1]
\State \textbf{Input:} Pseudo Glimpse Rationale $P_{gli}^{Pseudo}$, Pseudo Gaze Rationale $P_{gaz}^{Pseudo}$,
Situation $s = \{v, \mathcal{F}_v\}$ 
\State \textbf{Output:} Positive and Negative Glimpse and Gaze Rationales

\Function{Multi-round LLM-judgment and refinement}{$P_{gli}^{Pseudo}$, $P_{gaz}^{Pseudo}$, $s$}
    \State $rating \gets$ \Call{LLM-Judge}{$P_{gli/gaz}^{Pseudo}, s$}
    \While{$rating < N$}
        \State $P_{gli+/gaz+}^{refined} \gets$ \Call{Refine-Rationale}{$P_{{gli/gaz}}^{{Pseudo}}, s$}
        \State $rating \gets$ \Call{LLM-Judge}{$P_{{gli+/gaz+}}^{{refined}}, s$}
    \EndWhile
    \State \Return $P_{gli+}$, $P_{gaz+}$
\EndFunction

\Function{LLM-Judge}{$P, s$} - ``Single Answer
Grading''
    \State Please act as an impartial judge and evaluate the quality of the ``$P$''. Rate the $P$ that describes the given ``$s$'' on a score of 1 to 10, considering factors such as relevance, accuracy, detail...
    \State \Return $score$
\EndFunction

\Function{Refine-Rationale}{$P, s$}
    \State Refine the sentence based on the given pseudo ``$P$'' by incorporating relevant knowledge from the provided activity and/or entities words in the given ``$s$.'' Ensure the activity and/or entities in the sentence are present and clearly described.
    \State \Return $P^{{refined}}$
\EndFunction

\Function{GenerateNegativeRationale}{$P_{gli+/gaz+}, P_{gli/gaz}^{Pseudo}$}
    \State Generate a negative rationale based on the $P_{gli+/gaz+}$ and $P_{gli/gaz}^{Pseudo}$ by modifying the activity, entities, and attributes such as action, object, ..., and pattern with semantically similar...
    \State \Return $R_{{gli-}}$, $R_{{gaz-}}$
\EndFunction 
\end{algorithmic}
\end{algorithm}
\endgroup

In this process, we draw inspiration from the ``single answer grading'' judgment method~\cite{judging}, where the LLM-judge directly assigns a score to a rationale for describing the situation $s^b$  as shown in Algorithm~\ref{alg:llm-judge}. 
We use the same LLM to refine the rationales through multiple rounds of judgment and refinement if the assigned score is low (e.g., $<N=8$), ensuring that the generated rationales accurately describe the depicted situation. 
This process facilitates the accurate generation of positive glimpse $\mathbf{P}_{gli+}$ and gaze $\mathbf{P}_{gaz+}$ rationales. 
Additionally, we introduce a step where the LLM produces negative rationales $\mathbf{P}_{gli-}$ and $\mathbf{P}_{gaz-}$ by leveraging the general outputs of the MLLM. 
The negative rationale generation stays closely aligned with the positive text features but introduces variations in attribute information, helping the model better distinguish unseen and rare situations with negative distance loss (Eq.~\ref{eq5}).
These accurately refined rationales from large models~\cite{deepseek-r1, gemini1.5} provide rich semantic information, enabling the student model to acquire generalization knowledge~\cite{Distmultistage, distrecom, CMD-SE} during distillation process, thereby enhancing its situation recognition ability.

\vspace{-6pt}

\subsubsection{Multimodal Interactive Prompt Distillation Framework}
The Multimodal Interactive Prompt Distillation (MIPD) framework distills rich multimodal knowledge from generated rationales and visual information from the teacher MLLM model into the student model. 
This enhances the student’s generalization ability to recognize both activity and entities and improves its performance in rare and unseen situations.
To facilitate distillation, we introduce scene-aware and instance-perception soft prompts that capture holistic and perceptual visual representations from the MLLM. These prompts are interactively aligned with glimpse and gaze information from rationales, transferring multimodal knowledge to the student Ov-GSR model. Alignment is achieved with the Negative-Guided Multimodal Prompting Alignment (NMPA) module.

\textbf{Scene-aware and instance-perception prompts construction:} Scene-aware prompts function as a glimpse-based knowledge distiller, enabling the student model to absorb both holistic visual and glimpse semantic knowledge from the teacher model.
A straightforward approach to constructing the prompt is to leverage the recently advanced visual prompting technique~\cite{visionprompt, visionprompt2}, which attaches learnable prefixes to the input image patch and fine-tunes with supervised labels for improved performance. However, this method may introduce perturbations that affect feature extraction when the vision encoder is frozen and without direct optimization with ground truth, eventually impacting the distillation process. 

Hence, we construct scene-aware learnable visual prompts $\mathbf{P}_{sce} \in \mathbb{R}^{D \times \left( 2p(H + W - 2p) \right)}$ and append them to the edges of the visual features $\mathbf{X}_T$ extracted from the frozen encoder of the teacher MLLM, without affecting its feature extraction. These prompts absorb holistic visual knowledge from $\mathbf{X}_T$ and semantic cues from the glimpse rationale $\mathbf{P}{gli+}$, which are then distilled into the student model.

\begin{table*}[t]
\centering
\caption{ \label{tab:1} \textbf{Results (\%) of Ov-GSR methods on the Ov-SWiG dataset}, including three settings and five metrics evaluated on the dev and test set. The higher the number the better the performance. * denotes the model uses open-vocabulary settings~\cite{THID, hoiclip, CMD-SE}. Bold number represents highest accuracy.}
\vspace{-3mm}
\resizebox{1\textwidth}{!}{
\tiny
\begin{tabular}{l|ccccc|ccc|ccc}
\hline
\multicolumn{1}{c|}{\multirow{2}{*}{Models}} & \multicolumn{5}{c|}{Top-1-all}      & \multicolumn{3}{c|}{Top-1-rare}      & \multicolumn{3}{c}{Top-1-unseen}              \\
\multicolumn{1}{c|}{} & verb & value& val-all        & grnd & grnd-all       & verb & value       & grnd   & verb    & value        & grnd   \\ \hline
\multicolumn{12}{c}{Ov-GSR dev set} \\  \hline  
OpenSU*~\cite{openSU} & {36.28} & {30.03} & 18.86 & {20.27} & {6.35} 
& {23.40} & {18.05}  & {8.68} 
& {3.20} & {1.86}  & {1.00} \\
ClipSite*~\cite{clipsitu} & {38.60} & {31.49} & {20.27} & {21.03} & {7.08} 
& {24.70} & {18.46} & {10.00} 
& {3.60} & {2.08} & {1.43}\\
THID~\cite{THID} & {37.24} & {29.94} & {19.49} & {20.70} & {6.84} 
& {25.20} & {19.42} & {11.41} & {5.00} & {3.49} & {2.80} \\
CMD-SE~\cite{CMD-SE} & {39.04} & {32.67} & {20.64} & {21.35} & {7.29} 
& {27.40} & {21.70} & {12.73} & {6.40} & {4.05} & {3.05} \\
{MIPD (Ours)} & \bf{41.87} & \bf{34.29} & \bf{22.02} & \bf{23.29} & \bf{7.85} 
& \bf{29.10} & \bf{23.58} & \bf{14.65} & \bf{7.80}& \bf{4.73}& \bf{3.97} \\ 
\hline 
\multicolumn{12}{c}{Ov-GSR test set} \\ 
\hline  
OpenSU*~\cite{openSU} & {36.42} & {30.07} & {18.13} & {19.95} & {6.21} 
& {23.40} & {18.04} & {8.61} & {2.40} & {1.73} & {0.60}  \\
ClipSite*~\cite{clipsitu} & {38.64} & {31.51} & {20.15} & {20.84} & {6.75} 
& {24.60} & {18.50} & {9.24} & {3.00} & {1.86} & {1.20} \\
THID~\cite{THID} & {37.57} & {29.53} & {19.24} & 20.40 & {6.40} 
& {25.50} & {18.99} & {10.70} & {4.80} & {3.19} & {2.40} \\
CMD-SE~\cite{CMD-SE} & {39.17} & {32.69} & {20.29} & {20.97} & {7.01} 
& {26.80} & {20.29} & {11.48} & {6.00} & {3.45} & {2.57} \\
{MIPD (Ours)} & \bf{41.96} & \bf{34.11} & \bf{21.56} & \bf{22.86} & \bf{7.57} & \bf{28.30} & \bf{22.37} & \bf{13.59} & \bf{7.40}& \bf{4.08}& \bf{3.53} \\
\hline
\end{tabular}} 
\end{table*}

Instance-perception prompts serve as gaze-based knowledge distillers, enabling the student model to better understand entities with regional information. We construct learnable prompts $\mathbf{P}_{ins} \in \mathbb{R}^{D \times b_i \times H' \times W'}$ based on instance coordinates $b_i \in \mathbb{R}^4$, absorbing perceptual cues from the teacher $\mathbf{X}^T$ along with semantic $\mathbf{P}_{gaz+}$ information, where $H'$ and $W'$ are the height and width of each box. This prompt is distilled into the student model to improve entity awareness, which also bridges the gap between image-level pretrained MLLM and instance-level understanding of Ov-GSR.


\textbf{Negative-guided Multimodal Prompting Alignment:} 
We introduce an NMPA module to align positive glimpse rationales  $\mathbf{P}_{gli}$ with $\mathbf{P}_{sce}$ and gaze rationales $\mathbf{P}_{gaz}$ with $\mathbf{P}_{ins}$. 
This module integrates semantic insights with holistic and regional visual knowledge from the teacher model, facilitating generalized situation awareness in the student model during the distillation process. 
Furthermore, we employ a negative-guided prompting distance loss to ensure that the carefully crafted negative rationale remains close to the positive feature in the latent space~\cite{devilOv-detect, negativeloss}, preserving semantic similarity while improving situation discrimination. This allows the visual prompts and teacher features $\mathbf{X}_T$ to align positively with the positive rationales $\mathbf{P}_+$ and diverge from the negative rationales $\mathbf{P}_-$, enhancing situation awareness, which can be formulated as:
\begin{equation}\label{eq5} 
    \mathcal{L}_{neg} = - [\text{sim}(\mathbf{X}^{glimpse}_T, \mathbf{P}_{gli-}) + \text{sim}(\mathbf{X}^{gaze}_T, \mathbf{P}_{gaz-})]
\end{equation}
\begin{equation}\label{eq3}
    \mathbf{X}^{glimpse}_T = \delta_{glimpse}([\mathbf{X}_T + \mathbf{P}_{sce}]W^q, \mathbf{P}_{gli+}W^{kv})
\end{equation}
\begin{equation}\label{eq6}
    \mathbf{X}^{gaze}_T = \delta_{gaze}([\mathbf{X}_T + \mathbf{P}_{ins}]W^q,  \mathbf{P}_{gaz+}W^{kv})
\end{equation}
where $W^*$ is the projection parameters. ``sim'' denotes the cosine similarity function. In this work, we adopt simple yet effective cross-attention, denoted as $\delta_{glimpse}(\cdot)$ and $\delta_{gaze}(\cdot)$, to facilitate alignment between the learnable prompts and positive rationals. We further employ a negative guided distance loss $\mathcal{L}_{neg}$ to better correlate the positive and negative representations during the distillation process. 
NMPA aligns comprehensive multimodal knowledge from the teacher MLLM model with prompts and distills them into the student model, narrowing the gap between seen and unseen scenarios and reducing bias in rare situation predictions.

\begin{table}[t!]
\setlength{\tabcolsep}{3.8pt}
    \centering
     \caption{Comparison of our proposed MIPD with state-of-the-art methods on HICO-DET dataset. \cmark \ indicates the use of a pre-trained DETR~\cite{detr}, while \xmark \ means the model is trained without it and under settings similar to ours.} \vspace{-10pt}
    \begin{tabular}{l c c  c c}
        \hline
        {Method} & {Pretrained Detector}  & {Unseen} & {Seen} & {Full} \\
        \hline
        VCL~\cite{VCL} & \cmark &  10.06 & 24.28 & 21.43 \\
        ATL~\cite{ATL} & \cmark &  9.18 & 24.67 & 21.57 \\
        FCL~\cite{FCL} & \cmark &  13.16 & 24.23 & 22.01 \\
        GEN-VLKT~\cite{Gen-vlkt} & \cmark &  21.36 & 32.91 & 30.56 \\
        HOICLIP~\cite{hoiclip} & \cmark &  {23.48} & {34.47} & {32.26} \\
        DHD~\cite{DHD} & \cmark &  {23.32} & {30.09} & {28.53} \\
        \hline
        THID~\cite{THID}  & \xmark &  15.53 & {24.32} & {22.38} \\
        CMD-SE~\cite{CMD-SE} & \xmark &  {16.70} & 23.95 & 22.35 \\
        MIPD (Ours) & \xmark &  \bf{17.84} & \bf{25.45} & \bf{23.96} \\
        \hline
    \end{tabular}
    \label{tab:2} 
\end{table}


\subsection{Overall Training and Inference Process}

\vspace{-2pt}

During training, we generate text embeddings $t^v$ and $t^r$ using a frozen CLIP~\cite{clip} text encoder for activities and entities that correspond to the role classes in the situation set~\cite{THID, CMD-SE}.
Then, we compute the similarity between activity text embeddings and visual features using matrix multiplication, where the activity visual embeddings are defined as $\mathbf{\varepsilon}_s^{v} = \beta(\mathbf{X}_SW^v \cdot t^v)$. Here, $\mathbf{X}_S \in \mathbb{R}^{D \times H \times W}$ denotes the projected visual features from the frozen student model's CLIP vision encoder~\cite{clip}, and  $W^v$ is an additional projection layer serves as the activity head. The softmax function $\beta$ is applied for activity prediction.
Similarly, we calculate the similarity between entity embeddings with the role visual embeddings $\mathbf{\varepsilon}_s^{r} = \beta(\mathbf{X}_S^r\mathbf{W}^r \cdot t^r)$ with the additional multihead self-attention modules $\phi(\cdot)$ and projector as the role head, where  $\mathbf{X}_s^{r} = \phi([\mathbf{X}_S, \mathbf{\Bar{X}}^{v}_S]W)$. $\mathbf{\Bar{X}}^{v}_S$ represents the mean-pooled activity features, which are added to guide the prediction of roles. This is based on the constraint that a situation is only considered correct if the activity is accurately predicted for GSR~\cite{GSR, gsrformer}.
The classification loss for the \textbf{situation objective} is:
\begin{equation} 
    \mathcal{L}_{sit} = \mathcal{L}_{ce}(\mathbf{\varepsilon}_s^{v},v^b) + \sum_{\text{f}_r \in \mathcal{F}_v ^b} \mathcal{L}_{ce}(\mathbf{\varepsilon}_s^{r},\text{f}_r) 
\end{equation}
where $\mathcal{L}_{ce}$ denotes cross-entropy loss. \textbf{Distillation Objective:} With $\mathbf{X}_T^{glimpse}$ and $\mathbf{X}_T^{gaze}$ from the teacher model, we can distill their knowledge to student with the L1 loss functions as follow:
\begin{equation} 
    \mathcal{L}_{dis}  = \left| \mathbf{X}_T^{glimpse} - \mathbf{X}_S \right| 
    + \left| \mathbf{X}_T^{gaze} - \mathbf{X}_S^r \right|
\end{equation}
Additionally, the bounding box $\mathcal{L}_{box}$ optimization loss~\cite{GSR, coformer} is included for localization. The total loss can be computed as: $\mathcal{L} = \mathcal{L}_{neg}+\mathcal{L}_{sit}+\mathcal{L}_{dis} + \mathcal{L}_{box}$.

At the inference phase: the student model $S_{\text{model}}$ predicts the situation $\hat{s}$ with image $I$ input:
\begin{equation}
    \hat{s} = \{\hat{v}, \hat{\mathcal{F}}\} = \arg\max_{\{v, \mathcal{F}\} \in \mathcal{S}} P(\{v, \mathcal{F}\}|S_{\text{model}}(I;\theta_T)),
\end{equation}
where $s={\{v, \mathcal{F}\}}$ is a situation in $\mathcal{S}=\mathcal{S}^b \cup \mathcal{S}^u$ and $P(\cdot|\cdot)$ denotes the Bayesian posterior probability.


\section{Experiment}


\subsection{Experimental Settings} \label{setting}
\textbf{Benchmark dataset.} We evaluate our Ov-GSR approach on the newly split Ov-SWiG dataset, built upon SWiG~\cite{GSR}, containing 124,384 images split into 73,984 for training, and 25,200 each for development and test sets.
Each image is paired with three verb frames annotated by three different annotators. 
We use 500 images within dev and test set to evaluate open-vocabulary performance for 1,500 unseen situation pairs, where 10 unseen verbs are randomly picked from frequently used to rarely used for this evaluate. This resulting 67 entity annotation is not seen in the training set.
We select the last 20 rarely seen verbs, resulting in 3,000 rare situation pairs to evaluate the rare cases.
Ov-SWIG dataset has 504 verb categories, 190 semantic role types, and 9928 object categories, where each verb is associated with 1 to 6 semantic roles.
In addition, we further conduct experiments on the relevant HICO-DET dataset~\cite{HICO-det}. 
It consists of 600 interaction combinations, encompassing 117 human actions and 80 objects. Following~\cite{CMD-SE, THID}, we simulate a zero-shot detection setting by excluding 120 rare interactions from the full set of 600.

\textbf{Evaluation metrics.} 
We follow prior GSR works~\cite{GSR, gsrformer, coformer} and adopt five evaluation metrics to assess our method: (1) verb: activity prediction accuracy; (2) value: entity prediction accuracy per role; (3) val-all: entity prediction accuracy for the full semantic role set; (4) grnd: grounding (localization) accuracy per role; and (5) grnd-all: grounding accuracy across all semantic role set.
A grounding prediction is considered correct if its Intersection-over-Union (IoU) with the ground truth is $ \geq$ 0.5. Metrics are evaluated under three settings: Top-1-all, Top-1-rare, and Top-1-unseen. Semantic role predictions (e.g., val, grnd) are deemed incorrect if the activity prediction is incorrect.
We evaluate the performance of HICO-DET using mean Average Precision (mAP) that is the same as the existing methods ~\cite{HICO-det, CMD-SE}. An HOI triplet prediction is a true positive if the IoU between both the human and object bounding boxes exceeds 0.5, and the predicted interaction category is correct.

\textbf{Implementation Details.}
We utilize frozen CLIP~\cite{clip} (CLIP-ViT-L14) as the student model and InstrutBlip~\cite{instructblip} as the MLLM teacher to conduct experiment. All the dimensions of visual and text embeddings will project to D=512 in the experiment. The training learning rate of the proposed model is $10^{-4}$. We use AdamW Optimizer with a weight decay of $10^{-4}$, where $\beta_1 = 0.9$, and $\beta_2 = 0.999$. We train our proposed model for 10 epochs with a batch size of 32 on a single RTX3090 GPU, including the model analysis on figure~\ref{fig:tab9}.

\begin{table}[t!]
\setlength{\tabcolsep}{5.8pt}
\centering 
\caption{ \label{tab:3} \textbf{Results (\%) of close-set GSR methods on the original SWiG dataset}, evaluated on the the test set for top-1-all.} \vspace{-10pt}
\begin{tabular}{l|ccccc}
\hline
\multicolumn{1}{c|}{\multirow{2}{*}{Models}} & \multicolumn{5}{c}{Top-1-all}                 \\
\multicolumn{1}{c|}{} & verb & value & val-all        & grnd & grnd-all \\ 
\hline
\multicolumn{6}{c}{Close-set GSR test set} \\  \hline 
GSRTR~\cite{GSRTR}  & 40.63 & 32.15 & 19.28   & 25.49 & 10.10  \\
SituFormer~\cite{situformer}& 44.20 & 35.24 & 21.86   & 29.22 & 13.41    \\
CoFormer~\cite{coformer} & 44.66 & 35.98 & 22.22 & 29.05 & 12.21\\
GSRFormer~\cite{gsrformer} & {46.53} & {37.48} & {23.32} & {31.53} & {14.23} \\
OpenSU~\cite{openSU} & {50.10}&{41.20}&{26.56}&{34.27}&{15.70}\\
{ClipSitu}~\cite{clipsitu} & {58.19} & {47.23} & {29.73} & {40.01} & {15.03} \\ 
{MIPD (Ours)} & \bf{58.86} & \bf{49.33} & \bf{31.18} & \bf{41.78} & \bf{16.08} \\ 
\hline
\end{tabular} 
\end{table}

\subsection{Comparisons with existing methods}\vspace{-2pt}
Table \ref{tab:1} compares the performance of Multimodal Interactive Prompt Distillation (MIPD) on the Ov-SWIG benchmark with existing approaches on the dev and test set. These approaches include the grounded situation recognition methods OpenSU~\cite{openSU} and ClipSiTU~\cite{clipsitu} with an open-vocabulary classifier~\cite{hoiclip, CMD-SE}. Additionally, we compare MIPD with open-vocabulary HOI methods trained end-to-end under the same experimental settings, such as THID~\cite{THID} and CMD-SE~\cite{CMD-SE}, which leverage knowledge from large foundation models. These methods are re-implemented using a same setting to ours for a standardized and fair comparison. 
The 190 semantic roles~\cite{GSR} are seen all the time, so the value-all and grand-all matrices for roles are not included in rare and unseen scenarios. 
We observe that existing GSR models~\cite{openSU, clipsitu} primarily improve Top-1-all performance but struggle with rare and unseen cases, suggesting a prediction bias toward frequent and previously seen situations.
In contrast, open-vocabulary methods~\cite{THID, CMD-SE} demonstrate better performance in rare and unseen cases.
However, our proposed MIPD outperforms existing methods across top-1-all, rare, and unseen prediction settings. 
It improves unseen recognition and better identifies rare situations compared to previous approaches, demonstrating its strong generalization capabilities. 
We further apply our method to HOI detection and present the performance in Table~\ref{tab:2}, where our proposed approach outperforms other open-vocabulary HOI methods~\cite{THID, CMD-SE} under the same end-to-end training setting (without an additional pre-trained object detector~\cite{detr}) with ViT-B16, achieving superior unseen performance of 17.84\% and rarely seen performance of 25.45\%, demonstrating its effectiveness.
Moreover, we follow the setting of ClipSitu~\cite{clipsitu} and compare the performance of the MIPD with existing GSR approaches in Table \ref{tab:3} on the closed-set SWiG dataset~\cite{GSR}, the result further validates the effectiveness of our proposed method.
The improvement is partially attributed to better recognition of rare situations, as shown in Tables~\ref{tab:1} and~\ref{tab:2}, which contributes to overall GSR performance gains.
This further underscores the importance of distilling rich generalized multimodal knowledge from MLLMs to improve situation recognition.

\begin{table}[t!]
\setlength{\tabcolsep}{6.5pt}
  \centering
  \caption{The ablation results include comparisons with the baseline and direct distillation (w/ Dist) that use rationales.} \vspace{-8pt}
    \begin{tabular}{l|rrrrrr}
    \hline
    Method & \multicolumn{2}{c|}{All} & \multicolumn{2}{c|}{Rare} & \multicolumn{2}{c}{Unseen} \\
          & \multicolumn{1}{c}{verb} & \multicolumn{1}{c|}{value} & \multicolumn{1}{c}{verb} & \multicolumn{1}{c|}{value} & \multicolumn{1}{c}{verb} & \multicolumn{1}{c}{value} \\
    \hline
    Baseline &  36.78     &   29.44    &   22.70    &   17.91    &  2.80    & 1.85 \\
    w/ Dist &  38.76     &  31.54     &  25.80     & 20.33    &   5.60    & 3.47 \\
    w/ MIPD &   41.96    &  34.11     &   28.30   &   22.37    &   7.40    & 4.08 \\
    \hline
    \end{tabular}%
  \label{tab:4} 
\end{table}%

\begin{table}[t!]
\setlength{\tabcolsep}{3pt}
  \centering
  \caption{The ablation results analyze the impact of different prompts used in the model.} \vspace{-8pt}
    \begin{tabular}{cccc|cccccc}
    \hline
    \multicolumn{4}{c|}{Prompts}  & \multicolumn{2}{c|}{All} & \multicolumn{2}{c|}{Rare} & \multicolumn{2}{c}{Unseen} \\
    \multicolumn{1}{c}{$ \mathbf{P}_{sce}$} & \multicolumn{1}{c}{ $\mathbf{P}_{ins}$} & \multicolumn{1}{c}{ $\mathbf{P}_{gli}$} & \multicolumn{1}{c|}{ $\mathbf{P}_{gaz}$} & \multicolumn{1}{c}{verb} & \multicolumn{1}{c|}{value} & \multicolumn{1}{c}{verb} & \multicolumn{1}{c|}{value} & \multicolumn{1}{c}{verb} & \multicolumn{1}{c}{value} \\
    \hline
      \xmark    &  \xmark     &  \xmark     &  \xmark     &  36.78     &   29.44    &   22.70    &   17.91    &  2.80    & 1.85 \\
      \cmark    &  \cmark     &  \xmark     &  \xmark     &  38.28     &  30.66     &  24.20     & 18.33    &   3.40    & 2.13  \\
       \cmark    &   \xmark     &   \cmark     &   \xmark     &  40.41    &   31.70    &  27.70    &   20.69    &   6.80    & 3.86 \\
      \xmark  &    \cmark   &  \xmark     &   \cmark    &  37.36     &  30.98     &  23.80     & 19.33    &   3.20    & 2.95 \\
        \cmark  &    \cmark   &  \cmark     &  \cmark   &  41.96    &  34.11     &   28.30   &   22.37    &   7.40    & 4.08 \\
    \hline
    \end{tabular}%
  \label{tab:5} 
\end{table}%


\subsection{Ablation Studies}\vspace{-2pt}
We conduct an ablation study to analyze the impact of different model architectures and evaluate the effectiveness of Ov-GSR training. The experiments are conducted on the test set. The ablation study primarily highlights performance using the \underline{verb} and \underline{value} metrics, as they are most influenced by the model. 

Our method employs Multimodal Interactive Prompt Distillation (MIPD) to transfer knowledge from large models to a smaller model. Table~\ref{tab:4} presents the performance comparison between the baseline (without distillation), direct rationale distillation (w/ Dist), and our proposed approach (w/ MIPD) on the test set, illustrating the impact of MIPD on overall model performance. The results highlight the effectiveness of our distillation process, which improves overall situation recognition of verb and value metrics by around 5.2\% and 4.7\% compared to baseline, respectively.

In Table~\ref{tab:5}, we showcase the performance of using different prompts, highlighting its impact on model performance. We observe that even with the inclusion of visual-based only $\mathbf{P}_{sce}$ and $\mathbf{P}_{ins}$, the model shows improved situation awareness compared to the baseline. Incorporating $\mathbf{P}_{sce}$ and $\mathbf{P}_{gli}$ enhances scene-level activity understanding, leading to higher performance on the verb metric. Similarly, using $\mathbf{P}_{ins}$ and $\mathbf{P}_{gaz}$ improves entity recognition, resulting in better performance on the value metric compared to the base case.
We can see that incorporating all prompts enhances the model's ability to recognize rare and unseen situations by effectively aligning glimpse and gaze rationals with scene, and instance-level prompts.
The results show that these prompts effectively distill both semantic and visual knowledge into the student model, improving the Ov-GSR performance across seen, rare, and unseen situations.

\begin{table}[t!]
\setlength{\tabcolsep}{6.8pt}
  \centering
  \caption{The ablation results analyze the impact of different learnable visual prompts used in the model.}\vspace{-8pt}
    \begin{tabular}{l|cccccc}
    \hline
    \multicolumn{1}{l|}{Visual} & \multicolumn{2}{c|}{All} & \multicolumn{2}{c|}{Rare} & \multicolumn{2}{c}{Unseen} \\

    \multicolumn{1}{l|}{Prompts} & \multicolumn{1}{c}{verb} & \multicolumn{1}{c|}{value} & \multicolumn{1}{c}{verb} & \multicolumn{1}{c|}{value} & \multicolumn{1}{c}{verb} & \multicolumn{1}{c}{value} \\
    \hline
    baseline      &  36.78     &   29.44    &   22.70    &   17.91    &  2.80    & 1.85 \\
    Pad      &   39.50    &    31.95   &   27.10    &   19.85    &   6.20    &  3.52 \\
    Ours    &  41.96    &  34.11     &   28.30   &   22.37    &   7.40    & 4.08\\
    \hline
    \end{tabular}%
  \label{tab:6} 
\end{table}%

\begin{table}[t!]
\setlength{\tabcolsep}{6.5pt}
  \centering
  \caption{Ablation results showing the impact of using negative-guided distance loss ($\mathcal{L}_{neg}$) on model performance.}\vspace{-8pt}
    \begin{tabular}{c|cccccc}
    \hline
    \multicolumn{1}{l|}{Negative} & \multicolumn{2}{c|}{All} & \multicolumn{2}{c|}{Rare} & \multicolumn{2}{c}{Unseen} \\

    \multicolumn{1}{l|}{Rationals} & \multicolumn{1}{c}{verb} & \multicolumn{1}{c|}{value} & \multicolumn{1}{c}{verb} & \multicolumn{1}{c|}{value} & \multicolumn{1}{c}{verb} & \multicolumn{1}{c}{value} \\
    \hline
     \xmark     &  40.63     &   32.78    &    26.90   &   20.74    &   6.60    & 3.83 \\
      \cmark    &  41.96    &  34.11     &   28.30   &   22.37    &   7.40    & 4.08\\
    \hline
    \end{tabular}%
  \label{tab:7} 
\end{table}%

In Table~\ref{tab:6}, we illustrate the performance of visual prompting techniques that support the student model during the distillation process. Specifically, we compare the widely used PadPrompt~\cite{visionprompt} with our proposed scene-aware and instance-perception prompts. 
While PadPrompt improves knowledge transfer over the baseline, it was mostly designed for direct optimization with labels in a frozen model setting~\cite{visionprompt, VisualproTrack, Diversityprompting}, which may not be well-suited for distillation-based designs.
Our introduced prompts provide additional gains by enabling the student model to capture both holistic and regional visual-semantic information using the dense features from the MLLM, leading to improved recognition of seen, rare, and unseen situations as compared to padprompt.

Table~\ref{tab:7} presents an ablation study on the effect of using negative-guided distance loss ($\mathcal{L}_{neg}$) in model training. The results indicate that without incorporating negative rationales, model performance is lower across all evaluation settings. By introducing the negative-guided loss, the model achieves consistent improvements in both rare and unseen situations. These results demonstrate that encouraging the model to contrast informative negatives enhances its discriminative and generalization ability in our case, aligning with findings in prior works~\cite{devilOv-detect, negativeloss} that has similar concept.

Table~\ref{tab:8} presents the ablation results analyzing the impact of pseudo rationales and refined rationales using different judgment scores N. 
Based on experiments and observations, we set the maximum score to 8, as it offers comparable quality to scores of 9 or higher while requiring fewer reasoning rounds and lower cost.
The result show that models using MLLM-generated pseudo rationales perform suboptimally.
Refining rationales with a judgment score of 5 (D5) with Deepseek-r1~\cite{deepseek-r1} improves performance, with further gains observed at higher score 8 like D8 (Deepseek-r1) and G8 (geinimi-1.5-flash~\cite{gemini1.5}).
D8 performs better than G8 by generating more informative rationales, particularly for gaze rationales, which help Ov-GSR achieve better results. 
This experiment shows that higher-quality rationales, selected through JRG, can significantly enhance the student Ov-GSR model's ability to generalize, leading to better recognition of seen, rare, and unseen situations.


\begin{figure}[t!]
\centering
  \includegraphics[width=8.5cm, height=3.5cm]{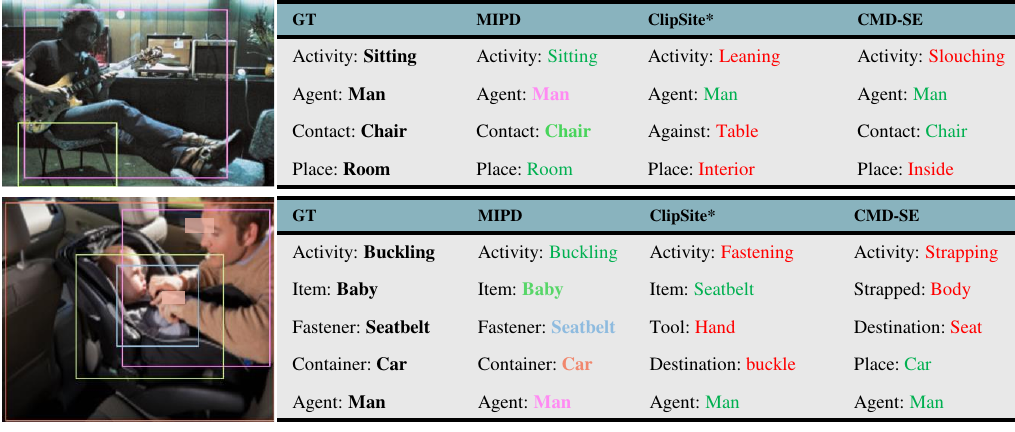} \vspace{-20pt}
  \caption{Examples of unseen situations (top) and rare situations (bottom). Green is correct predictions, red indicates incorrect ones, and bold colored text highlights our correct predictions with grounding. }
  \label{fig:4} 
  \vspace{-18pt}
\end{figure}

\subsection{Qualitative results}\vspace{-2pt}
The figure~\ref{fig:4} presents a qualitative comparison of our proposed MIPD framework with ClipSitu~\cite{clipsitu} and CMD-SE~\cite{CMD-SE} on unseen (top) and rare (bottom) situations. In the unseen example, where the ground truth activity is Sitting, MIPD accurately predicts the activity along with all corresponding entities and their roles. In contrast, ClipSitu and CMD-SE misclassify the activity as Leaning and Slouching, respectively, and fail to identify several semantic roles correctly. 
In the rare example involving the activity Buckling, MIPD again aligns well with the ground truth, accurately detecting entities like Baby, Seatbelt, Car, and Man. In contrast, both ClipSitu and CMD-SE misidentify the activity and incorrectly label several semantic roles.
These examples highlight MIPD’s ability to generalize beyond seen data, demonstrating its robustness in handling both unseen and rare grounded situation recognition.

\begin{table}[t!]
\setlength{\tabcolsep}{4.2pt}
  \centering
  \caption{Ablation results analyzing the impact of rationales and different scores used for refining rationales.}\vspace{-8pt}
    \begin{tabular}{lc|cccccc}
    \hline
    Rationales & \multicolumn{1}{c|}{Judge} & \multicolumn{2}{c|}{All} & \multicolumn{2}{c|}{Rare} & \multicolumn{2}{c}{Unseen} \\
          & \multicolumn{1}{c|}{score} & \multicolumn{1}{c}{verb} & \multicolumn{1}{c|}{value} & \multicolumn{1}{c}{verb} & \multicolumn{1}{c|}{value} & \multicolumn{1}{c}{verb} & \multicolumn{1}{c}{value} \\
    \hline
    Pesudo &   \xmark  &  {37.15} & {29.31}      &    25.10 & 18.22   &  4.40     & 3.25\\
    Refined & D5     &   40.26    &   32.83    &  26.70    &   20.73    &  6.40      & 3.72 \\
    Refined & G8     &  41.40    &  33.35     &   27.60   &   21.64    &   7.00    & 3.88  \\
    Refined & D8     &  41.96    &  34.11     &   28.30   &   22.37    &   7.40    & 4.08 \\
    \hline
    \end{tabular}%
  \label{tab:8} \vspace{-10pt}
\end{table}%


\section{Conclusion}\vspace{-2pt}
In this work, we tackle the novel and challenging problem of Ov-GSR by focusing on distilling knowledge from large models into smaller models to improve generalization to rare and unseen situations, and achieving better GSR performance.
We introduce the Multimodal Interactive Prompt Distillation (MIPD) framework, which distills rich semantic and visual knowledge to the student model by leveraging glimpse and gaze rationales aligned with scene-aware and instance-perception prompts. 
This alignment is achieved by the Negative-Guided Multimodal Prompting Alignment (NMPA) module, which allows the prompts to encapsulate holistic and perception-level knowledge.
Extensive experiments on Ov-SWiG and HICO-DET demonstrate that MIPD achieves state-of-the-art performance, confirming its effectiveness.

\bibliographystyle{ACM-Reference-Format}
\bibliography{sample-base}

\end{document}